\pdfoutput=1
\documentclass[]{llncs}
\usepackage[utf8]{inputenc}
\usepackage{multirow}
\usepackage{numprint}
\usepackage{amsmath}
\usepackage{booktabs}
\usepackage{verbatim}
\usepackage{color}

\newcommand{\beq}{\begin{equation}}
\newcommand{\eeq}{\end{equation}}

\renewcommand{\vec}[1]{\mathbf{#1}}

\newcommand{\mypar}[1]{\paragraph{\normalfont\textbf{#1}}}

\newcommand{\mr}[1]{\mathrm{#1}}

\newcommand{\htrain}{\vec{h}^{\mr{train}}}
\newcommand{\htest}{\vec{h}^{\mr{test}}}
\newcommand{\xtrain}{\vec{x}^{\mr{train}}}
\newcommand{\xtest}{\vec{x}^{\mr{test}}}

\title{Out-of-Distribution Detection for Skin Lesion Images with Deep Isolation Forest}

\author{Xuan Li\inst{1} \and
Yuchen Lu\inst{2} \and
Christian Desrosiers\inst{3} \and
Xue Liu\inst{1} \\
xuan.li2@mcgill.ca,
yuchen.lu@umontreal.ca,
christian.desrosiers@etsmtl.ca,
xue.liu@mcgill.ca}

\institute{McGill University \and
Universite de Montrea \and
ETS Montreal
}

\date{ }

\usepackage{graphicx}

\begin{document}

\maketitle

\begin{abstract}
In this paper, we study the problem of out-of-distribution detection in skin disease images. Publicly available medical datasets normally have a limited number of lesion classes (e.g. HAM10000 has 8 lesion classes). However, there exists a few thousands of clinically identified diseases. Hence, it is important if lesions not in the training data can be differentiated. Toward this goal, we propose DeepIF, a non-parametric Isolation Forest based approach combined with deep convolutional networks. We conduct comprehensive experiments to compare our DeepIF with three baseline models. Results demonstrate state-of-the-art performance of our proposed approach on the task of detecting abnormal skin lesions.

\end{abstract}

\section{Introduction}

Deep learning models such as the convolution neural networks (CNN) have shown outstanding potential in dermatology for skin cancer classification~\cite{esteva2017dermatologist,zhang2018towards,han2018deep}. 
However, the diversity of real life skin disease still hinder the application of automatic differential diagnosis to real life. E.g., the well-known HAM10000 dataset \cite{tschandl2018ham10000} contains eight different skin lesion classes in its training set. This is quite small compared to the actual number of known skin lesion types and subtypes, which can be in the thousands \cite{esteva2017dermatologist}. Hence, it is important to have methods that can make use of the limited amount of disease types in existing datasets to detect the unseen diseases. This is the problem of Out-Of-Distribution (OOD) detection, or abnormality detection. Recent work~\cite{lee2018simple} proposes a simple but effective OOD detection framework. They model a class conditional Guassian distribution on the final feature of any pre-trained neural network, and they use Mahalanobis-distance-based metric to compute the abnormality score. However, skin lesions, even within the same class, are known to have huge intra-class difference. As a result, we argue that a uni-modal Gaussian distribution might not be expressive enough to capture the distribution of representation, which is shown in our paper. 


To address this limitation, we propose to replace the simple Guassian estimation with a powerful non-parametric method Isolation Forest (IF)~\cite{liu2008isolation}. Unlike traditional anomaly detection techniques, IF does not require normal profiling nor assuming a distribution family for normal samples. IF is designed based on the intuition that, abnormal samples are few and different, and as a result, they can be easily classified by a decision tree with fewer splits~\cite{liu2008isolation}.
In this work, we propose to use IF on the features computed by a pre-trained deep CNN to detect OOD images of skin lesions, and hence the name \emph{DeepIF}. Our contributions are as follows:
\begin{itemize}\setlength\itemsep{3pt}
    \item We propose DeepIF as a modification to the existing OOD framework~\cite{lee2018simple} to take into account the huge intra-class diversity of skin disease image.  
    \item Our experiment on HAM10000 dataset~\cite{tschandl2018ham10000} shows that DeepIF outperform the existing baselines on OOD detection, and it provides a 20\% detection rate improvement compared to the metric based on simple Gaussian~\cite{lee2018simple}.
    \item 
    We present a comprehensive analysis of hidden representations from different convolutional layers. Results show that the last convolutional layer has the most expressive representations among most of the diseases. 
\end{itemize}

\section{Related Works}\label{sec:related-works}
In recent years, a broad range of approaches based on deep learning have been proposed for this problem. \cite{hendrycks2016baseline} introduce a simple heuristic by applying a threshold on the softmax probability of the predicted class. The ODIN approach, proposed by Liang et al. \cite{liang2018enhancing}, uses softmax temperature scaling and adversarial input perturbation to make the softmax scores of in-distribution and out-of-distribution examples better separated. Based on the assumption that features computed by a pre-trained network follow a class-conditional Gaussian distribution, Lee et al.~\cite{lee2018simple} use the Mahalanobis distance in the predicted class distribution to detect OOD and adversarial samples. Our method can be viewed as a non-parametric model extension on the above framework to take into account the high complexity of medical images like skin disease.
\par
In \cite{devries2018learning}, Devries et al. use an auxiliary loss function to generate a confidence score in another branch. The extra loss function encourages the network to identify examples for which its prediction is unsure. Vyas et al. \cite{vyas2018out} train an ensemble of classifiers in a self-supervised manner, considering a random subset of training examples as OOD data and the rest as in-distribution data. A margin-based loss is proposed to impose a given margin between the mean entropy of OOD and in-distribution samples. In \cite{masana2018metric}, Masana et al. use metric learning to derive an embedding space where samples from the same in–distribution class form clusters that are separated from other in–distribution classes and OOD samples. \cite{ouardini2019towards} propose to use transfer learning as a general abnormality detection for medical images.
\cite{ren2019likelihood} propose using the likelihood ratio between the output probability of two deep networks, the first one modeling in-distribution data and the second capturing background statistics, as measure of normality.
\par
While all these approaches require modifying the original training algorithm of the model, our method is more flexible as it only needs a pre-trained network and can use a black-box algorithm for training. In addition, these studies focus on natural images and, as shown in our experiments, do not work well on skin lesion images which have less inter-class variability. So far, only a few works have investigated OOD detection for this type of image. Pacheco et al.~\cite{pacheco2019skin} use the mean Shannon entropy of the softmax output for correctly classified and misclassified validation examples to detect outliers, yielding a 11.45\% OOD detection rate for the ISIC 2019 dataset. In a different approach, Lu et al.~\cite{lu2018anomaly} consider the likelihood of a variational autoencoder (VAE) to identify OOD skin lesion images. Different from these approaches, our method does not presume any distribution for the anomaly class. As we will empirically demonstrate, this makes our OOD method more robust. 


\section{Method}\label{sec:method}
    
\begin{figure}[t!]
\centering
\includegraphics[width=0.9\textwidth]{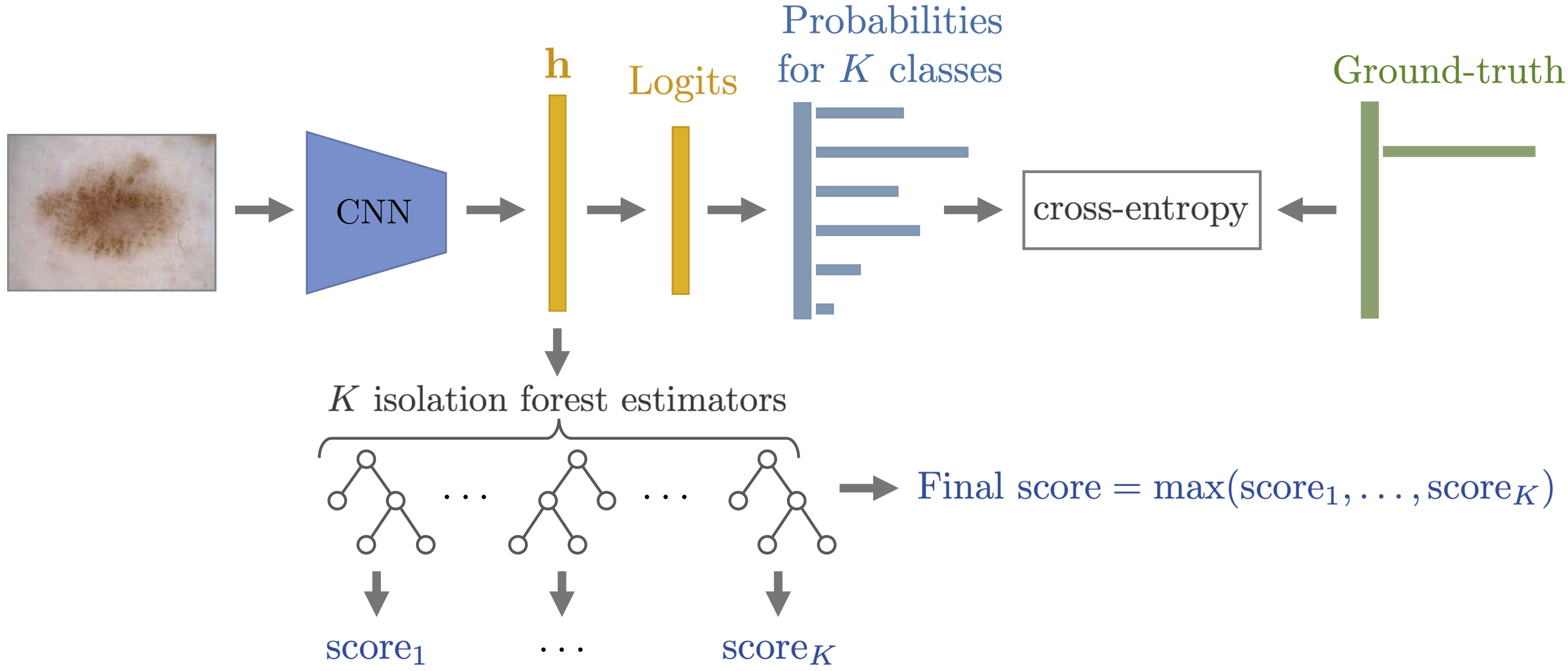}
\caption{Proposed DeepIF method for detecting OOD skin lesion images.}
\label{fig:training} 
\end{figure}

\mypar{Isolation Forest} Isolation Forest (IF) is an anomaly detection algorithm built on the idea of decision tree ensembling. Each decision tree is constructed by the data points in the training set. At each node of a tree, select a random feature from a subset of features (the proportion of the size of subset is $N_{f}$). A random value between the minimum and maximum values of that feature is chosen to make a split at that node. We construct a total of $N_{e}$ decision trees.
\par
For a given isolation forest $IF$ and the test data $\xtest$, we calculate the normality as:
\begin{equation}\label{eqn:isolation_forest_score}
    IF(\xtest) \ = \ -2^{-\frac{E[P_{e}]}{P_{\mr{avg}}}} \, + \, 0.5.
\end{equation}
where $P_{e}$ is the number of tree nodes (i.e., path length) traversed by $\xtest$ from the root node to the terminal leaf node on the $e$-th decision tree, and we take its average across all trees in $IF$. $P_{\mr{avg}}$ is the average path length for training data. We refer to the original paper \cite{liu2008isolation} for detailed information. The intuition is that anomaly data points have extreme values on certain features, such that they can be easily isolated and have shorter paths. Thus $IF(\xtest)$ would be small if $\xtest$ is an OOD data.

\mypar{OOD Detection Framework} An arbitrary CNN $f(\cdot)$ is pre-trained to predict the $K$ normal classes of the training data. The parameters of $f$ are then fixed when training finishes. Afterwards, training examples $\xtrain$ are fed into $f$ to obtain their hidden representation $\htrain$ from the last convolutional layer. Lee et al~\cite{lee2018simple} calculate the class mean and covariance as class-conditional Gaussian distributions based on the $\htrain$. For OOD detection, they extract the $\htest$ from $\xtest$ and calculate the Mahalanobis distance of each class, and assign $\xtest$ the shortest distance as the final anomaly score.

\mypar{Deep Isolation Forest (DeepIF)} Our DeepIF shares the same idea for extracting $\vec{h}$ from a pre-trained CNN (see Fig~\ref{fig:training}). Different from their distance-based approach, we construct models $IF_1, IF_2, ..., IF_K$ for each class. Then our final normality score is computed as 
$
\max(IF_1(\htest),IF_2(\htest),...,IF_K(\htest)).
$

\section{Experiments}\label{sec:experiments}

\mypar{Data and setup} The data we use is from the HAM10000 \cite{tschandl2018ham10000} training set which contains 25,331 images with 8 classes: Melanoma (MEL), Melanocytic nevus (NV), Basal cell carcinoma (BCC), Actinic keratosis (AK), Benign keratosis (BKL), Dermatofibroma (DF), Vascular lesion (VASC), Squamous cell carcinoma (SCC). For each experiment, we hold out 1 class as an Anomaly Class, which we refer to as an \emph{OOD set}. For each remaining class, a 90\% - 10\% split is made for the training and validation sets. We treat the validation set as \emph{in-distribution set}. Since HAM10000~\cite{tschandl2018ham10000} contains 8 classes, we conduct 8 experiments with a single class being treated as the Anomaly class and the rest 7 are normal classes in each experiment. 

\mypar{Pre-trained CNN} We train a skin lesion classification network with a standard approach: an image is feed into a ResNet152 \cite{he2016deep} to get the predictions for each class. Cross-entropy loss is calculated and back-propagated to the network. SGD is adopted to optimize the network with a learning rate of 1e-4. We train the network 200 epochs with a batch size of 32. In the training stage, one class is held out to be treated as an anomaly class. Once the training procedure finishes, the parameters of the network is fixed through the rest of the procedures.

For constructing the $\mr{IF}$  models, we set $N_e$ to be 200, and $N_f$ to be 1.0. Final scores for in-distribution and OOD sets are stored separately for evaluation.

\mypar{Baselines}
Our first baseline is to compare with the originally \emph{Mahalanobis-distance} baseline using the implementation from \cite{deepMahalanobisDetector}. 
We also compare to other strong baselines that beyond our framework. We compare to a \emph{Confidence Score} baseline~\cite{devries2018learning}, which learns to predict the confidence score. We use  the implementation from \cite{confidenceEstimation} but with the same network architecture as our DeepIF. Finally we compare with the \emph{VAE} baseline~\cite{lu2018anomaly} by measuring the negated reconstruction score.

\mypar{Evaluation Metrics} We adopt the same metrics as in other studies on OOD detection \cite{devries2018learning,lee2018simple,liang2018enhancing}: area under the ROC curve (\emph{AUROC}); area under the precision recall curve where in-distribution is specified as the positive (\emph{AUPR\,in}); area under the precision recall curve where OOD is specified as the positive (\emph{AUPR\,out}); true negative rate (TNR) when the true positive rate is as high as 95\% (\emph{TNR95TPR}). In the latter, the TNR is computed as TN/(TN+FP), where TN is the number of true negative and FP the number of false positives. We also show the classification accuracy on the validation dataset.

\section{Results}\label{sec:results}

The results are shown in Table~\ref{tab:ood_experiment}. We can first find that the confidence-based baseline would decrease the classification performance on validation data, with 4\% mean accuracy drop than the other methods. We believe that learning to predict confidence would add extra requirement to the training process which might hurt the performance of the main task, and therefore an OOD framework that does not touch the training procedure like ours has the advantage to preserve the model performance. 

DeepIF easily beat the Mahalanobis baseline, which confirms our hypothesis that medical images like skin lesion are too complex to be properly modelled by a uni-mode Gaussian even on the representation space. Our method also beat the VAE baseline, and VAE is known to be a very distribution modelling for high-dimensional data. We believe that this results show the potential of non-parametric OOD detection that does not depend on normal profiling~\cite{liu2008isolation}. The strongest baseline is the confidence score. DeepIF is better except in one metric (AUPR in), but DeepIF preserves the model accuracy. 

\begin{table}[ht!]
    \centering
    \caption{Results for OOD Experiment on HAM10000. We take take one class of images out of dataset as OOD set and only train on the rest of them.}
    \label{tab:ood_experiment}
    \npdecimalsign{.}
    \nprounddigits{4}
 \renewcommand{\arraystretch}{.75}    
\setlength{\tabcolsep}{6pt}    
\begin{footnotesize}
\begin{tabular}{l|l|ccccc}
\toprule
\textbf{OOD} & \multicolumn{1}{c|}{\multirow{2}{*}{\textbf{Method}}} & \multirow{2}{*}{\textbf{AUROC}} & \textbf{AUPR}   & \textbf{AUPR} & \textbf{TNR at} & \textbf{Val. Acc \%} \\

\textbf{Set} & & & \textbf{in} & \textbf{out} & \textbf{95\% TPR} & \\
 \midrule
\multirow{4}{*}{MEL}
& DeepIF     & \textbf{0.6918}  & \textbf{0.6856}     & \textbf{0.6909}    & \textbf{0.1969}  & \multirow{3}{*}{\textbf{93.3}}\\
& Mahalanobis & 0.6108  & 0.5797     & 0.6073    & 0.1186  &   \\
& VAE      & 0.5653  & 0.5619     & 0.5301    & 0.0411  &  \\ \cmidrule{3-7}
& Confidence   & 0.6248  & 0.6536     & 0.5555    & 0.0249 &  89.5\\
\midrule

\multirow{4}{*}{NV} 
& DeepIF     & \textbf{0.6311}  & \textbf{0.6513}     & \textbf{0.5969}    & \textbf{0.0894}  & \multirow{3}{*}{\textbf{90.7}}\\
& Mahalanobis & 0.5537  & 0.5564     & 0.5525    & 0.0807   & \\
& VAE       & 0.5545  & 0.5606     & 0.5201    & 0.0362  & \\ \cmidrule{3-7}
& Confidence   & 0.4375  & 0.5011     & 0.4301    & 0.0041  & 84.1\\
 \midrule
 
\multirow{4}{*}{BCC} 
& DeepIF   & 0.7539  & 0.6878     & 0.7503    & 0.2724   & \multirow{3}{*}{\textbf{89.0}}\\ 
& Mahalanobis & 0.5702  & 0.5785     & 0.5347    & 0.0464   & \\
& VAE    & 0.5292  & 0.5324     & 0.5109    & 0.0453    & \\ \cmidrule{3-7}
& Confidence   & \textbf{0.8236}  & \textbf{0.8325}     & \textbf{0.7921}    & \textbf{0.2996}  & 85.3 \\
 \midrule
\multirow{4}{*}{AK} 
& DeepIF     & 0.6942  & 0.6271     & 0.6879    & 0.1693  & \multirow{3}{*}{\textbf{90.3}}\\ 
& Mahalanobis & 0.5509  & 0.5304     & 0.5398    & 0.0741   & \\
& VAE     & 0.5151  & 0.5195     & 0.4938    & 0.0316  & \\ \cmidrule{3-7}
& Confidence     & \textbf{0.7908}  & \textbf{0.8136}     & \textbf{0.7416}    & \textbf{0.1929} & 86.4\\
 \midrule
 
\multirow{4}{*}{BKL} 
& DeepIF   & 0.6991  & 0.6743     & \textbf{0.6847}    & \textbf{0.1738}  &  \multirow{3}{*}{\textbf{91.2}}\\ 
& Mahalanobis  & 0.6126  & 0.6031     & 0.5790    & 0.0729  &  \\
& VAE     & 0.5151  & 0.5195     & 0.4938    & 0.0316  &  \\ \cmidrule{3-7}
& Confidence     & \textbf{0.7384}  & \textbf{0.7611}     & 0.6698    & 0.1032 & 87.2 \\
 \midrule
 
\multirow{4}{*}{DF} 
& DeepIF & \textbf{0.7462}  & 0.7108     & \textbf{0.7302}    & \textbf{0.2676}   & \multirow{3}{*}{\textbf{88.9}} \\ 
& Mahalanobis  & 0.5443  & 0.5409     & 0.5188    & 0.0584  &   \\
& VAE     & 0.5230  & 0.5468     & 0.5040    & 0.0375  &   \\ \cmidrule{3-7}
& Confidence      & 0.6972  & \textbf{0.7389}     & 0.6279    & 0.0858 & 84.6\\
 \midrule
\multirow{4}{*}{VASC} 
& DeepIF    & \textbf{0.7483}  & \textbf{0.7480}     & \textbf{0.7040}    & \textbf{0.1635}   & \multirow{3}{*}{\textbf{89.6}}\\ %
  & Mahalanobis & 0.5985  & 0.6229     & 0.5467    & 0.0466 &  \\
  & VAE      & 0.5159  & 0.5490     & 0.4808    & 0.0221  & \\ \cmidrule{3-7}
& Confidence      & 0.4813  & 0.5579     & 0.4489    & 0.0118 & 84.7\\
 \midrule
\multirow{4}{*}{SCC} 
& DeepIF    & 0.7612  & 0.7105     & 0.7523    & 0.2573  &\multirow{3}{*}{\textbf{89.2}} \\ 
& Mahalanobis & 0.5758  & 0.5705     & 0.5342    & 0.0397   &\\
& VAE   & 0.5336  & 0.5444     & 0.5096    & 0.0404   &\\ \cmidrule{3-7}
& Confidence      & \textbf{0.8324}  & \textbf{0.8505}     & \textbf{0.7858}    & \textbf{0.2682} & 86.1\\ 
\midrule
\midrule
 \multirow{4}{*}{\textbf{Mean}}
& DeepIF & \textbf{0.7136} & 0.6841 & \textbf{0.6985} & \textbf{0.1979} & \multirow{3}{*}{\textbf{90.3}}\\
& Mahalanobis & 0.5771 & 0.5728 & 0.5516 & 0.0672 &\\
& VAE & 0.5315 & 0.5418 & 0.5054 & 0.0357 & \\ \cmidrule{3-7}
& Confidence   & 0.6783 & \textbf{0.7137} & 0.6315 & 0.1238 & 86.1 \\
\bottomrule
\end{tabular}
\end{footnotesize}
\end{table}



\begin{figure}[t!]
\centering
\begin{tabular}{cc}
\includegraphics[width=.48\textwidth]{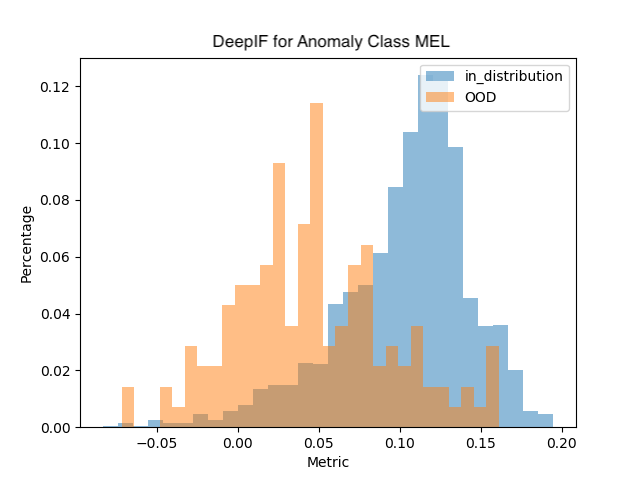} & 
\includegraphics[width=.48\textwidth]{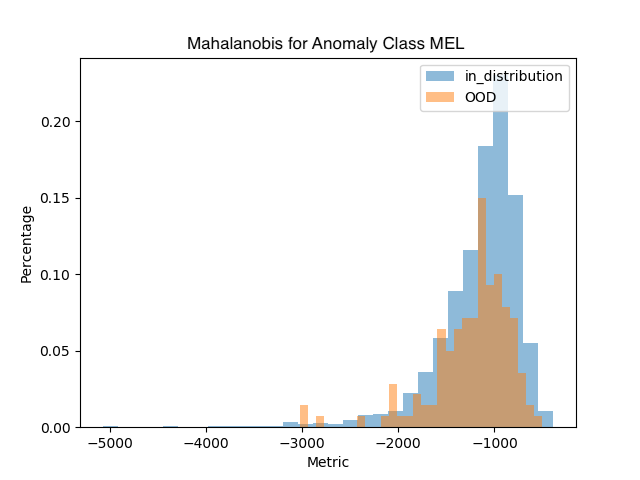} \\
\end{tabular}
\caption{Comparison of normality score distribution between DeepIF and Mahalanobis baseline. The OOD set is MEL.}
\label{fig:distribution}
\end{figure}

\begin{figure}[ht!]
\centering
\setlength{\tabcolsep}{10pt} 
\begin{tabular}{cc}
\includegraphics[width=.42\textwidth]{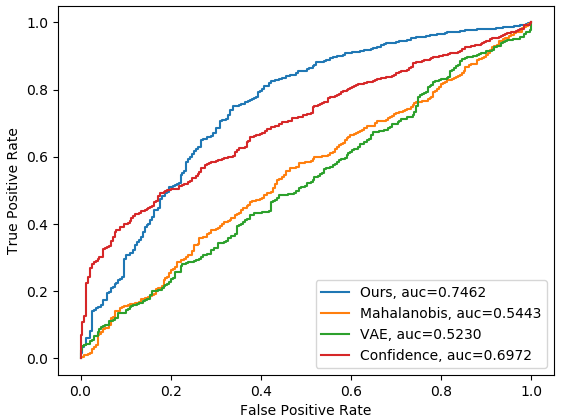} & 
\includegraphics[width=.42\textwidth]{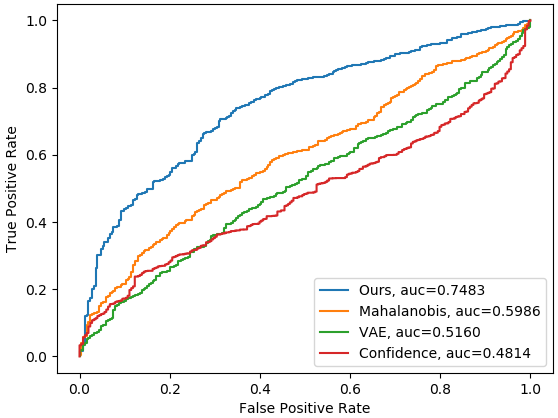} \\
\end{tabular}
\caption{ROC curves for OOD set to be BKL (left) and OOD set to be DF (right). DeepIF yields a better ROC curve compared with the other 3 approaches. }
\label{fig:roccurves}
\end{figure}

We plot in Fig~\ref{fig:distribution} the histograms of normality scores for in- and out-distribution data point between Mahalanobis baseline and DeepIF with MEL as the OOD set. It can be observed that DeepIF scores lead to a better separation of in-distribution and OOD examples, which explains our method's better ability in differentiating those two datasets. We also plot in Fig~\ref{fig:roccurves} the ROC curves with OOD set to be BKL and DF.  

We analyze the effect of using the representation from different layers. Our default choice is to use the last layer $f_{-1}$. We evaluate the performance of DeepIF from $f_{-2}$ to $f_{-4}$ as well. The result is shown in Table.~\ref{tab:layercomparison}. We find that, with the exception of NV, the performance of DeepIF with shallower features is worse than using deep features. This highlights the importance of semantic information captured in deeper layers for OOD detection. 

\begin{table}[ht!]
    \centering
    \caption{Result of DeepIF using features from different layers of the pretrained network. }
    \label{tab:layercomparison}
  \setlength{\tabcolsep}{6pt}    
  \renewcommand{\arraystretch}{.8}
  \begin{small}
\begin{tabular}{c|c|cccc}
 \toprule
 \textbf{OOD} & \multirow{2}{*}{\textbf{Layer}} & \multirow{2}{*}{\textbf{AUROC}} & \textbf{AUPR}   & \textbf{AUPR} & \textbf{TNR at} \\
 \textbf{set} & & & \textbf{in} & \textbf{out} & \textbf{95\% TPR}\\
 \midrule
 \multirow{4}{*}{MEL} & $f_{-1}$ & \textbf{0.6918}  & \textbf{0.6856}     & \textbf{0.6909}    & \textbf{0.1969} \\
                             & $f_{-2}$ & 0.6240 & 0.6323 & 0.6081 & 0.1109 \\
                             & $f_{-3}$ & 0.6001 & 0.5628 & 0.5921 & 0.1071 \\
                             & $f_{-4}$ & 0.5600 & 0.5265 & 0.5653 & 0.0977 \\
\midrule
 \multirow{4}{*}{NV} & $f_{-1}$ & 0.6311  & \textbf{0.6513}     & 0.5969    & 0.0894 \\
                             & $f_{-2}$ & 0.4891 & 0.4879 & 0.4941 & 0.0506 \\
                             & $f_{-3}$ & 0.6599 & 0.6329 & 0.6461 & 0.1114 \\
                             & $f_{-4}$ & \textbf{0.6628} & 0.6215 & \textbf{0.6642} & \textbf{0.1908} \\
\midrule
 \multirow{4}{*}{BCC} & $f_{-1}$ & \textbf{0.7539}  & \textbf{0.6878}     & \textbf{0.7503}    & \textbf{0.2724} \\
                             & $f_{-2}$ & 0.6244 & 0.6655 & 0.5865 & 0.0718 \\
                             & $f_{-3}$ & 0.5485 & 0.5484 & 0.5315 & 0.0596 \\
                             & $f_{-4}$ & 0.5296 & 0.5150 & 0.5229 & 0.0465 \\
\midrule
 \multirow{4}{*}{AK} & $f_{-1}$ & \textbf{0.6942}  & 0.6271     & \textbf{0.6879}    & \textbf{0.1693} \\
                             & $f_{-2}$ & 0.6652 & \textbf{0.7217} & 0.6028 & 0.0734 \\
                             & $f_{-3}$ & 0.5742 & 0.5992 & 0.5247 & 0.0352 \\
                             & $f_{-4}$ & 0.5494 & 0.5809 & 0.5113 & 0.0415 \\
\midrule
 \multirow{4}{*}{BKL} & $f_{-2}$ & \textbf{0.6991}  & \textbf{0.6743}     & \textbf{0.6847}    & \textbf{0.1738}  \\
                             & $f_{-2}$ & 0.5494 & 0.5809 & 0.5113 & 0.0415 \\
                             & $f_{-3}$ & 0.4920 & 0.4969 & 0.4891 & 0.0437 \\
                             & $f_{-4}$ & 0.4600 & 0.4718 & 0.4754 & 0.0420 \\
\midrule
 \multirow{4}{*}{DF} & $f_{-1}$ & \textbf{0.7462}  & \textbf{0.7108}  & \textbf{0.7302}  & \textbf{0.2676} \\
                             & $f_{-2}$ & 0.5116 & 0.5516 & 0.4886 & 0.0365 \\
                             & $f_{-3}$ & 0.4600 & 0.4813 & 0.4532 & 0.0134 \\
                             & $f_{-4}$ & 0.4521 & 0.4753 & 0.4518 & 0.0324 \\
\midrule
 \multirow{4}{*}{VASC} & $f_{-1}$ &\textbf{0.7483}  & \textbf{0.7480}   & \textbf{0.7040}  & \textbf{0.1635} \\
                             & $f_{-2}$ & 0.4888 & 0.5338 & 0.4940 & 0.0607 \\
                             & $f_{-3}$ & 0.5295 & 0.5394 & 0.5180 & 0.0731 \\
                             & $f_{-4}$ & 0.5432 & 0.5249 & 0.5554 & 0.1182 \\
\midrule
 \multirow{4}{*}{SCC} & $f_{-1}$ & \textbf{0.7612}  & \textbf{0.7105} & \textbf{0.7523}  & \textbf{0.2573}  \\
                             & $f_{-2}$ & 0.6575 & 0.6869 & 0.6159 & 0.0927 \\
                             & $f_{-3}$ & 0.5518 & 0.5461 & 0.5267 & 0.0417 \\
                             & $f_{-4}$ & 0.4774 & 0.4781 & 0.4904 & 0.0468 \\
                            
 \bottomrule
\end{tabular}
\end{small}
\end{table}

\section{Discussion and Conclusion}

In this paper, we studied the problem of OOD detection with a non-parametric approach on the HAM10000 \cite{tschandl2018ham10000} skin lesion dataset. We proposed a simple framework by adopting a pre-trained CNN and Isolation Forest models. Our experiments showed our approach to achieve state-of-the-art performance for differentiating in-distribution and OOD data.

We demonstrated the usefulness of our proposed DeepIF, method on a skin lesion dataset. To further validate our method, we aim to cover a broader range of medical image datasets where there exists huge intra-class diversity, for instance, Diabetic Retinopathy, CT, and MRI datasets. Moreover, while our DeepIF focuses on image data, our method can be easily transferred to other non-image data, such as electric medical records data, or time sequence data including electroencephalogram (EEG) and electrocardiogram (ECG). In future work, we would also like to compare test method with more non-parametric algorithms such as Dirichlet Process Mixture Model (DPMM)~\cite{blei2006variational} or a self-organizing network~\cite{marsland2002self}.

\newpage
\bibliographystyle{splncs04}
\bibliography{references}

\begin{thebibliography}{10}
\providecommand{\url}[1]{\texttt{#1}}
\providecommand{\urlprefix}{URL }
\providecommand{\doi}[1]{https://doi.org/#1}

\bibitem{blei2006variational}
Blei, D.M., Jordan, M.I., et~al.: Variational inference for dirichlet process
  mixtures. Bayesian analysis  \textbf{1}(1),  121--143 (2006)

\bibitem{confidenceEstimation}
DeVries, T.: Learning confidence for out-of-distribution detection in neural
  networks. \url{https://github.com/uoguelph-mlrg/confidence\_estimation}
  (2018)

\bibitem{devries2018learning}
DeVries, T., Taylor, G.W.: Learning confidence for out-of-distribution
  detection in neural networks. arXiv preprint arXiv:1802.04865  (2018)

\bibitem{esteva2017dermatologist}
Esteva, A., Kuprel, B., Novoa, R.A., Ko, J., Swetter, S.M., Blau, H.M., Thrun,
  S.: Dermatologist-level classification of skin cancer with deep neural
  networks. Nature  \textbf{542}(7639),  115--118 (2017)

\bibitem{han2018deep}
Han, S.S., Park, G.H., Lim, W., Kim, M.S., Im~Na, J., Park, I., Chang, S.E.:
  Deep neural networks show an equivalent and often superior performance to
  dermatologists in onychomycosis diagnosis: Automatic construction of
  onychomycosis datasets by region-based convolutional deep neural network.
  PloS one  \textbf{13}(1) (2018)

\bibitem{he2016deep}
He, K., Zhang, X., Ren, S., Sun, J.: Deep residual learning for image
  recognition. In: Proceedings of the IEEE conference on computer vision and
  pattern recognition. pp. 770--778 (2016)

\bibitem{hendrycks2016baseline}
Hendrycks, D., Gimpel, K.: A baseline for detecting misclassified and
  out-of-distribution examples in neural networks. arXiv preprint
  arXiv:1610.02136  (2016)

\bibitem{deepMahalanobisDetector}
Lee, K.: A simple unified framework for detecting out-of-distribution samples
  and adversarial attacks.
  \url{https://github.com/pokaxpoka/deep\_Mahalanobis\_detector} (2019)

\bibitem{lee2018simple}
Lee, K., Lee, K., Lee, H., Shin, J.: A simple unified framework for detecting
  out-of-distribution samples and adversarial attacks. In: Advances in Neural
  Information Processing Systems. pp. 7167--7177 (2018)

\bibitem{liang2018enhancing}
Liang, S., Li, Y., Srikant, R.: Enhancing the reliability of
  out-of-distribution image detection in neural networks. In: 6th International
  Conference on Learning Representations, ICLR 2018 (2018)

\bibitem{liu2008isolation}
Liu, F.T., Ting, K.M., Zhou, Z.H.: Isolation forest. In: 2008 Eighth IEEE
  International Conference on Data Mining. pp. 413--422. IEEE (2008)

\bibitem{lu2018anomaly}
Lu, Y., Xu, P.: Anomaly detection for skin disease images using variational
  autoencoder. arXiv preprint arXiv:1807.01349  (2018)

\bibitem{marsland2002self}
Marsland, S., Shapiro, J., Nehmzow, U.: A self-organising network that grows
  when required. Neural networks  \textbf{15}(8-9),  1041--1058 (2002)

\bibitem{masana2018metric}
Masana, M., Ruiz, I., Serrat, J., van~de Weijer, J., Lopez, A.M.: Metric
  learning for novelty and anomaly detection. arXiv preprint arXiv:1808.05492
  (2018)

\bibitem{ouardini2019towards}
Ouardini, K., Yang, H., Unnikrishnan, B., Romain, M., Garcin, C., Zenati, H.,
  Campbell, J.P., Chiang, M.F., Kalpathy-Cramer, J., Chandrasekhar, V., et~al.:
  Towards practical unsupervised anomaly detection on retinal images. In:
  Domain Adaptation and Representation Transfer and Medical Image Learning with
  Less Labels and Imperfect Data, pp. 225--234. Springer (2019)

\bibitem{pacheco2019skin}
Pacheco, A.G., Ali, A.R., Trappenberg, T.: Skin cancer detection based on deep
  learning and entropy to detect outlier samples. arXiv preprint
  arXiv:1909.04525  (2019)

\bibitem{ren2019likelihood}
Ren, J., Liu, P.J., Fertig, E., Snoek, J., Poplin, R., Depristo, M., Dillon,
  J., Lakshminarayanan, B.: Likelihood ratios for out-of-distribution
  detection. In: Advances in Neural Information Processing Systems. pp.
  14680--14691 (2019)

\bibitem{tschandl2018ham10000}
Tschandl, P., Rosendahl, C., Kittler, H.: The ham10000 dataset, a large
  collection of multi-source dermatoscopic images of common pigmented skin
  lesions. Scientific data  \textbf{5},  180161 (2018)

\bibitem{vyas2018out}
Vyas, A., Jammalamadaka, N., Zhu, X., Das, D., Kaul, B., Willke, T.L.:
  Out-of-distribution detection using an ensemble of self supervised leave-out
  classifiers. In: Proceedings of the European Conference on Computer Vision
  (ECCV). pp. 550--564 (2018)

\bibitem{zhang2018towards}
Zhang, X., Wang, S., Liu, J., Tao, C.: Towards improving diagnosis of skin
  diseases by combining deep neural network and human knowledge. BMC medical
  informatics and decision making  \textbf{18}(2), ~59 (2018)

\end{thebibliography}
\end{document}